\documentclass{sigchi}

\CopyrightYear{2021}
\setcopyright{acmlicensed}
\doi{https://doi.org/10.1145/3313831.XXXXXXX}
\conferenceinfo{HRI'21, Workshop "Exploring Applications for Autonomous Non-Verbal Human-Robot Interactions"}{March  8, 2021}

\usepackage{balance}       
\usepackage{graphics}      
\usepackage[T1]{fontenc}   
\usepackage{txfonts}
\usepackage{mathptmx}
\usepackage[pdflang={en-US},pdftex]{hyperref}
\usepackage{color}
\usepackage{booktabs}
\usepackage{textcomp}

\usepackage{microtype}        
\usepackage{ccicons}          

\usepackage{todonotes}

\def\plaintitle{It's your turn! - A collaborative human-robot pick-and-place scenario in a virtual industrial setting}

\def\emptyauthor{}
\def\plainkeywords{human-cobot interaction; pick-and-place actions; non-verbal signalling; virtual reality}

\makeatletter
\def\url@leostyle{%
  \@ifundefined{selectfont}{
    \def\UrlFont{\sf}
  }{
    \def\UrlFont{\small\bf\ttfamily}
  }}
\makeatother
\def\pprw{8.5in}
\def\pprh{11in}

\definecolor{linkColor}{RGB}{6,125,233}
\hypersetup{%
  pdftitle={\plaintitle},
  pdfauthor={\emptyauthor},
  pdfkeywords={\plainkeywords},
  pdfdisplaydoctitle=true, 
  bookmarksnumbered,
  pdfstartview={FitH},
  colorlinks,
  citecolor=black,
  filecolor=black,
  linkcolor=black,
  urlcolor=linkColor,
  breaklinks=true,
  hypertexnames=false
}

\begin{document}

\title{\plaintitle}

\numberofauthors{3}
\author{%
  \alignauthor{Leave Authors Anonymous\\
    \affaddr{for Submission}\\
    \affaddr{City, Country}\\
    \email{e-mail address}}\\
  \alignauthor{Leave Authors Anonymous\\
    \affaddr{for Submission}\\
    \affaddr{City, Country}\\
    \email{e-mail address}}\\
  \alignauthor{Leave Authors Anonymous\\
    \affaddr{for Submission}\\
    \affaddr{City, Country}\\
    \email{e-mail address}}\\
}

\author{%
  \alignauthor{Brigitte Krenn\\Tim Reinboth\\Stephanie Gross\\
    \affaddr{Austrian Research Institute for Artificial Intelligence}\\
    \affaddr{Vienna, Austria}\\
    \email{firstname.lastname@ofai.at}}\\
  \alignauthor{Christine Busch\\Martina Mara\\Kathrin Meyer\\
    \affaddr{LIT Robopsychology Lab, Johannes Kepler University}\\
    \affaddr{Linz, Austria}\\
    \email{firstname.lastname@jku.at}}\\
    \alignauthor{Michael Heiml\\Thomas Layer-Wagner\\
    \affaddr{Polycular e.U.}\\
    \affaddr{Hallein, Austria}\\
    \email{firstname.lastname@polycular.com}}\\
}
\maketitle

\begin{abstract}
In human-robot collaborative interaction scenarios, nonverbal communication plays an important role. Both, signals sent by a human collaborator need to be identified and interpreted by the robotic system, and the signals sent by the robot need to be identified and interpreted by the human. 
In this paper, we focus on the latter. We implemented on an industrial robot in a VR environment nonverbal behavior signalling the user that it is now their turn to proceed with a pick-and-place task. The signals were presented in four different test conditions: no signal, robot arm gesture, light signal, combination of robot arm gesture and light signal. Test conditions were presented in two rounds to the participants. The qualitative analysis was conducted with focus on (i) potential signals in human behaviour indicating why some participants immediately took over from the robot whereas others needed more time to explore, (ii) human reactions after the nonverbal signal of the robot, and (iii) whether participants showed different behaviours in the different test conditions. 
We could not identify potential signals why some participants were immediately successful and others not. There was a bandwidth of behaviors after the robot stopped working, e.g. participants rearranged the objects, looked at the robot or the object, or gestured the robot to proceed. We found evidence that robot deictic gestures are helpful for the human to correctly interpret what to do next. Moreover, there was a strong tendency that humans interpreted the light signal projected on the robot's gripper as a request to give the object in focus to the robot. Whereas a robot's pointing gesture at the object was a strong trigger for the humans to look at the object. 

\end{abstract}





\printccsdesc

\section{Introduction and Background Work}

As collaborative robots (cobots) become more prevalent in industries, investigating the interactions between humans and cobots becomes relevant, especially as there is strong evidence, that humans respond socially to computers \cite{Nass2000}. In this context, the investigation of robots' nonverbal communicative signals is of particular importance to foster user acceptance and enhance user experience as well as to facilitate successful joint task completion. 

In this paper, we qualitatively investigate human reactions to a "it's your turn" signal from an industrial robot in virtual reality. In this human-robot-collaboration experiment, a UR10 robot is signalling to a human during a pick-and-place task that it is now their turn to put small boxes from a conveyor belt in a cardboard box. 
In our research, we focus on nonverbal interaction, due to the unreliability of speech in task descriptions. In situated tasks, verbal descriptions of objects and actions are often imprecise and partially incorrect \cite{Gross2017,Gross2016}. Although not problematic for humans, this raises a challenge for automatic processing in robots. In addition, problems of automatic speech recognition systems in typically noisy industrial working environments speak against verbal interactions.

In general, nonverbal communication is the "unspoken dialogue" that creates shared meaning in social interactions \cite{Burgoon2016}. There are several categorizations of nonverbal communication in distinct, although socially interrelated modes, e.g. in kinesics, proxemics, haptics, chronemics, vocalics, and presentation \cite{Jones2013}. In the present paper, we focus on kinesics, a highly articulated form of nonverbal communication, specifically arm gestures. Saunderson et al. \cite{Saunderson2019} categorize kinesics-based robotics research into (i) arm gestures, (ii) body and head movements, (iii) eye gaze, and (iv) facial expressions.

El Zaatari et al. \cite{elZaatari2019} mention four types of interaction in industrial human-cobot collaboration: independent -- a worker and a cobot operate on separate work pieces; simultaneous -- a worker and a cobot operate on separate processes on the same work piece at the same time; sequential -- a worker and a cobot perform sequential manufacturing processes on the same work piece; supportive -- a worker and a cobot work towards the same process on the same work piece interactively. The signal "it's your turn" is potentially relevant for both sequential and supportive collaboration.


Studies have shown that robot arm gestures are influential in improving the response time in interactive tasks. In a cooperation task, Riek et al. \cite{Riek2010} investigated how humans might react to three robot arm gestures: beckoning, giving, shaking hands. The gestures were displayed by BERTI, a humanoid robot, manipulating the gesture style (abruptly or smoothly) and the gesture orientation (front or side). They were presented to the participants via videos, who then had to pick an action based on what the participant thought the robot is asking him or her to do. Results showed that people react to abrupt gestures more quickly than smooth ones and front-oriented gestures more quickly than side oriented gestures.

Quintero et al. \cite{Quintero2015} investigated pointing gestures in human-human and human-robot interaction while collaboratively performing manipulation tasks. Feedback via predefined "yes" and "no" confirmation gestures allowed the robot (a seven-degrees-of-freedom robot arm) or the human to verify that the right object was selected. Their results showed 28\% misinterpretations of robot pointing gestures by humans, 2\% misinterpretations of human pointing gestures by the robotic system and 10\% misinterpretations in human-human interaction. 

Sheikholeslami et al. \cite{Sheikholeslami2017} explored the efficacy of nonverbal communication capabilities of a robotic manipulator in a collaborative car door assembly task. First, the authors observed the type of hand configurations that humans use to nonverbally instruct another person. Then they examined how well human gestures with frequently used hand configurations are understood by the recipients. Subsequently they implemented the most human-recognized human hand configurations on a seven-degrees-of-freedom robotic manipulator to investigate the efficacy of human-inspired hand poses on a three-fingered gripper. In two video-based online surveys, human and robot gestures were presented to participants. Out of the 14 selected and implemented gestures, recognition rate of human gestures was greater than 90\%. 10 implemented gestures were relatively well recognized (recognition rate > 60\%) by the participants.
However, four gestures yielded lower recognition rates (left, right, swap, confirm), possibly due to the angle of the arm relative to the assembly object, the complexity of the task or physical limitations of the robot gripper such as lacking a thumb.

We implemented our experimental setting in virtual reality (VR), as it is a seminal technology offering several advantages not only for human-robot-interaction research \cite{Oyekan2019,Peters2018} but also for training courses and education  \cite{Perez2019,Burghardt2020}. 
For trainings, the safety aspect, the high degree of immersion and the thus resulting effectivity of trainings for low cost are emphasized by \cite{Perez2019} and \cite{Burghardt2020}. Oyekan et al. \cite{Oyekan2019} argue that due to high velocity and massive force of industrial robots, VR is an effective surrounding to investigate human reactions towards predictable and unpredictable robotic movements, and to develop strategies and solutions based on these reactions. Peters et al. \cite{Peters2018} emphasize the advantages of less expensive and time consuming HRI experiments in VR.

In the presented paper, we qualitatively investigated human reactive behaviours to a cobot displaying arm (beckoning, pointing) and light gestures as turn-giving signals. 

In general, the idea of focusing on sequences where the human does not take the parcel and put it into the box immediately after the robot stopped working, is that there may be more to learn about participants' behaviours during the task from instances in which participants were not immediately successful, rather than from those in which participants quickly succeeded in the task. The eye-gaze, motion capture and video recordings each provide data about what participants did, as they, presumably, tried to figure out what was going on, when the robotic arm stopped working. Whereas immediate success provides little data about how participants approached the situation, the recordings of all other trials provide rich data about what participants did in response to the robot's inaction. 
However, even in a relatively constrained environment such as the one reported her, human behaviour remains difficult to interpret. Even more elusive is why a particular human acted a certain way, meaning what exactly motivated the action and what that action was intended to achieve. It is instructive to bear in mind, that this way of thinking about human action deals with voluntary action in particular, while involuntary behaviours present further challenges of their own. That is why, rather than trying to infer intentions and chains of thought into the minds of the participants, we first turned our attention to what visibly happened.

\section{VR Game Scenario}

\begin{figure}
  \centering
  \includegraphics[width=0.9\columnwidth]{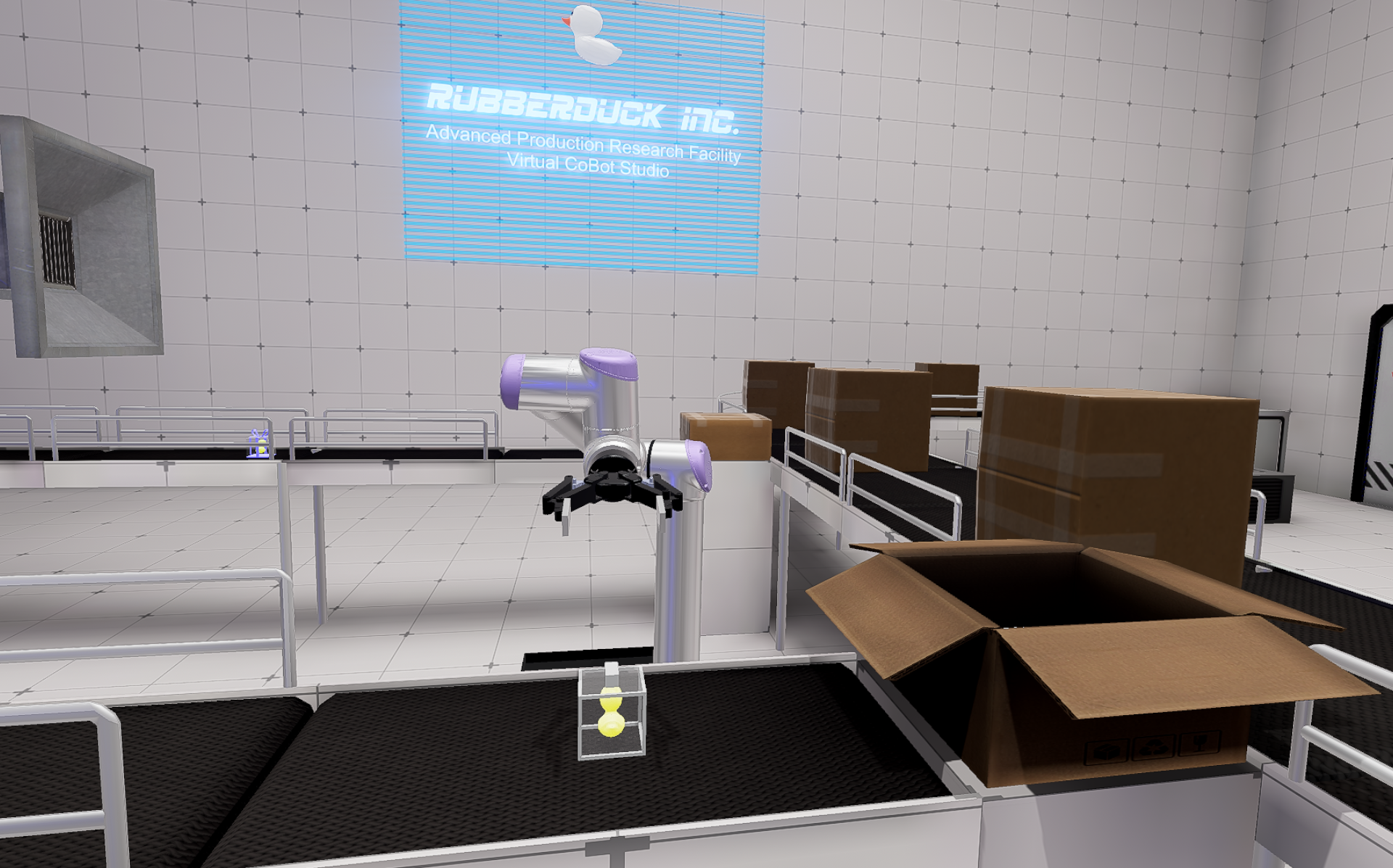}
  \caption{Screenshot: Person's view on the packaging hall.}~\label{fig:figure_sceneview}
\end{figure}

The game plays in a virtual rubber duck factory. Human and robot stand in front of each other separated by a conveyor belt. The robot is a digital twin, a 1:1 replica of the CHIMERA\footnote{\url{https://www.joanneum.at/en/robotics/infrastructure/mobile-manipulation}} mobile manipulator comprising a UR10 collaborative robot arm with a two-finger gripper mounted on a MiR100 mobile platform, having the same capabilities and looks as in real life. The conveyor belt transports parcels containing rubber ducks which need to be taken from the belt and put into a cardboard box which is positioned next to the robot. The task of the human is to watch the scene and help the robot when needed. 
Apart from her/his face to face position with the robot, the human has a 360 degree view of the factory hall, seeing the individual parcels coming from one side and the cardboard boxes filled with the duck parcels being transported away by a second conveyor belt. See Figure \ref{fig:figure_sceneview} for a view on the scenario from the human's perspective. Equipped with an optical hand tracking module on the VR headset, the human is able to take the parcel from the conveyor belt and to put it into the cardboard box next to the robot. S/he is also able to reorganize the parcels in the box. 

The typical game scenario is that the robot is working and the human is watching, i.e., the robot takes parcel per parcel from the conveyor belt and puts them into the box. At some point, the robot stops working and it is the human's turn to take action. The goal of the experimental setup is to test whether the human understands that it is now her/his turn and that s/he should do the robot's work. Apart from the situation where the robot arm stops with its gripper above the parcel (control condition), two types of signals were explored: (i) the robot arm produces human-like pointing -- first at the human, then at the parcel. The signal is meant to indicate that it is now the human's turn to pick up the parcel from the conveyor belt and to put it into the cardboard box; (ii) a light signal, i.e., a projection of downwards running arrows on the gripper. For more details on the experimental setup and the conditions tested see the following section.

\section{Experimental Setup}

To explore the interpretability of different motion- and light-based signals, we conducted a mixed design experiment in which participants collaborated on a packaging task with a mobile sensitive manipulator in a virtual reality mini game. This mini game is part of the research-based VR game CoBot Studio\footnote{For more information, see \url{https://www.jku.at/en/lit-robopsychology-lab/research/projects/cobot-studio/}}, taking place in a virtual rubber duck factory. 

\subsection{Manipulated Variables}
We manipulated one between-subjects factor (“it’s your turn” signal variant presented by the robot) and one within-subjects factor (number of rubber ducks presented per round). The signal variant was manipulated in the following four levels: 
\begin{itemize}\itemsep0pt
    \item Variant 1 (V1, control group): gripper stops above parcel with no further signaling.
    \item Variant 2 (V2): arm gestures, i.e., the robot arm points at the person, then at the parcel containing the rubber duck.
    \item Variant 3 (V3): gripper stops above parcel and demonstrates light signals – blue arrows running down the gripper.
    \item Variant 4 (V4): combination of robot arm gestures and light signal.
\end{itemize}

Participants collaborated with the robot in two game rounds which were completed in randomized order and varied regarding the number of rubber duck parcels being packaged by the robot before the participant was supposed to take over. The robot took either three or two rubber duck parcels from a conveyor belt and put them in a cardboard box, before signaling the participants to continue (or before simply stopping in the control condition). The rationale for choosing this setup was to assess potential
carry-over effects between game rounds.

\subsection{Procedure \& Task}
Once they had been welcomed, participants signed their agreement to the experimental procedure and filled out a questionnaire. Next, they received an introduction to the VR game CoBot Studio as well as to the VR equipment by a research assistant. Afterwards, participants put on the VR equipment which was then calibrated. If the participants indicated that they felt comfortable, an interactive tutorial sequence was started which helped participants to get accustomed to the VR environment. The packaging task took place as one of three different mini games within the CoBot Studio VR game that the participants subsequently played in random order. When starting the packaging task mini game, participants received an introduction and were informed about their task by a disembodied instructor. To account for as much variance in behavior as possible to the received signal cue, task information was kept short: The participants were only informed that they should watch the robot during its work and assist if needed.

After participants indicated that they were ready to start, the first mini game round started and proceeded as previously outlined. In each game round, the participant had 60 seconds to grab the rubber duck parcel after the robot indicated her/him to take over. If the first 60 seconds were exceeded (= failure), the participant received an information that the task had not been completed correctly. The game round was then terminated. If the participant grabbed the rubber duck parcel within the 60 second countdown, s/he was given another 60 seconds to put the parcel in the shipping carton (= success) before the game round was terminated. After completing two game rounds, participants received an outro and continued with a different mini game. Upon finishing the CoBot Studio VR game, they filled out another questionnaire, were thanked and rewarded with a miniature rubber duck. 

\section{Sample}
A total of n = 133 participants were recruited at an Austrian University and during a research exhibition. However, due to technical problems, the video files of nine participants were either not utilizable at all or defective regarding one of the game rounds. Hence, the analyses for round 1 and 2 were conducted on a data set of 124 participants respectively. Regarding their gender and age, the participant groups slightly differed in the two game rounds. In round 1, 55 (44\%) were female, 68 (55\%) were male and one person selected the option “other”.  The mean age of participants in round 1 was M = 31.18 (SD = 9.90), with a range from 14 to 65 years. In round 2, 54 (44\%) were female, 69 (56\%) were male and one person selected the option “other”. Their mean age was M = 31.14 years (SD = 9.91), again ranging from 14 to 65 years. 
The participant groups did not vary regarding their prior experience in working with industrial robots and their experience with VR headsets. 21 (17\%) stated that they had previously worked with an industrial robot, while 100 (81\%) had not previously worked with an industrial robot and 3 (2\%) stated that they were unsure. Moreover, 35 (28\%) had never used a VR headset, 50 (40\%) had once used a VR headset, and 39 (32\%) were regularly using VR headsets.

\section{Qualitative Study}

\begin{figure*}
  \centering
  \includegraphics[width=1.75\columnwidth]{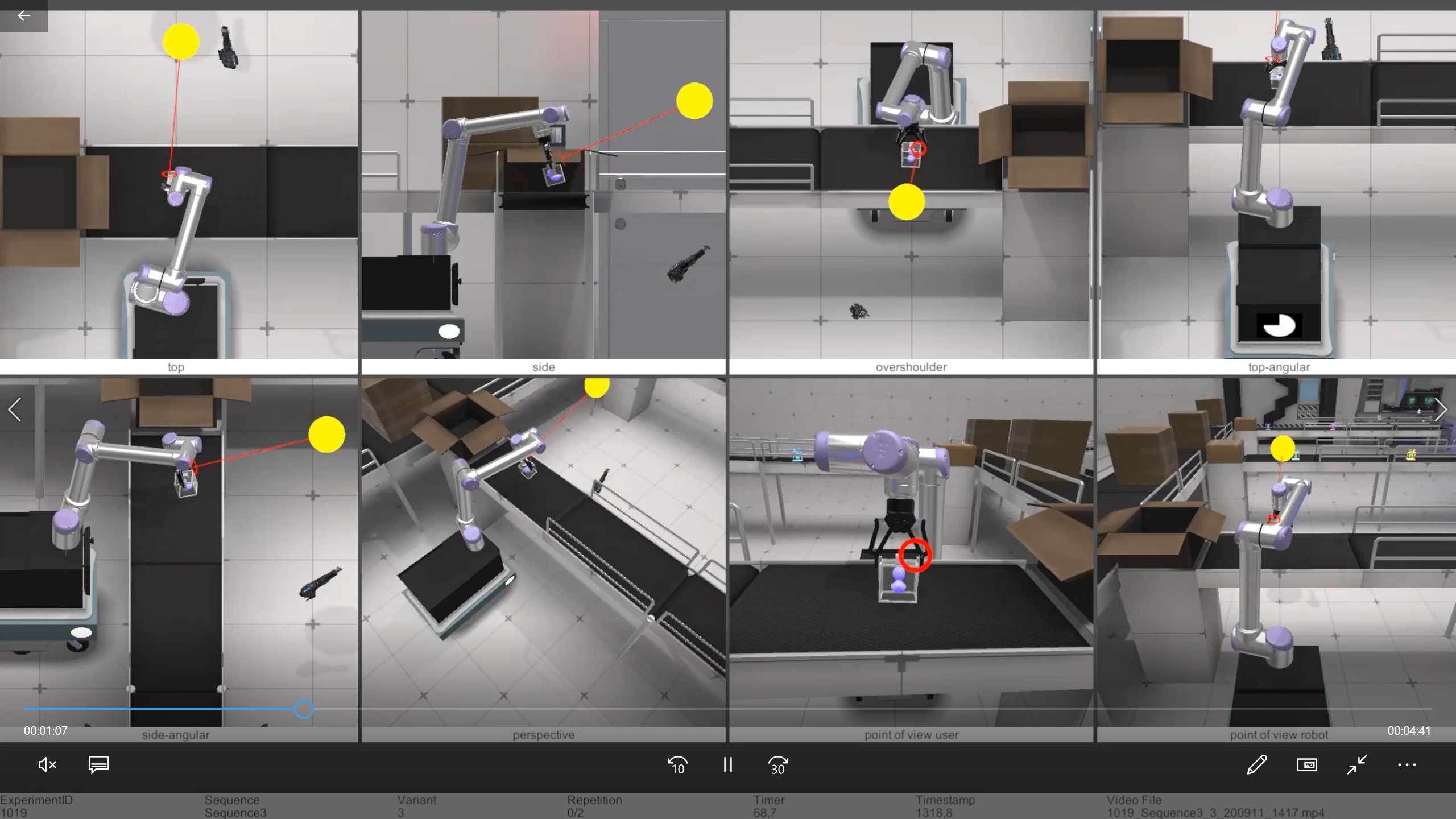}
  \caption{Screenshot from the recorded VR video, showing various views on an ongoing robot-user interaction; "point of view user" is the primary view for our analyses as it shows the human first person view.}~\label{fig:figure_allVR}
\end{figure*}

For each participant and experimental variant (V1-V4) a video was recorded in VR covering the following 8 perspectives (see Figure \ref{fig:figure_allVR} from left to right): top -- a bird's eye view on robot and user; side -- view from the left side; overshoulder -- from over the head of the user; top-angular -- from behind the robot, low to high; side-angular -- from the left side, elevated; perspective -- from the left side, diagonal and elevated; point of view user -- first person perspective human; point of view robot -- from behind the robot at human head level. The yellow dot represents the head of the user, the red line indicates the eye gaze coming from the eye tracker built into the VR headset, and the red circle indicates the focus point of the human gaze. The black device is the control stick the user holds in her/his left or right hand. The user's respective other hand is only visible in the "point of view user" video whenever the hand is in the user's field of vision. This way, we can also see when the human grabs a parcel and what s/he does with it.

In the qualitative analysis, we were interested in (i) potential signals in the human behaviour that could function as indicators for distinguishing immediately successful users from those who needed more time or failed to stand in for the robot and put the parcel into the cardboard box; 
(ii) human reactions after the robot had stopped working; 
(iii) whether the participants showed different behaviours in reaction to the test conditions V1 to V4. As regards (i), we could not find any indicators in the users' overt behaviour that would give hint on who would more or less quickly grasp the task. 

\begin{figure}[hbt]
  \centering
  \includegraphics[width=0.7\columnwidth]{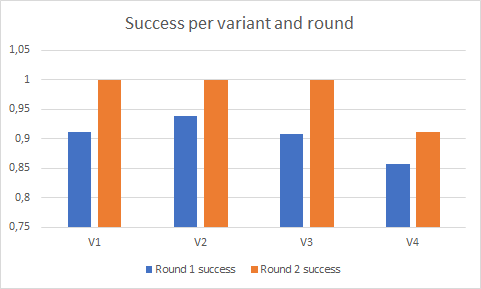}
  \caption{Comparison of successes rates (\%) for the 4 study conditions V1-V4. We distinguish overall success within  the 1\textsuperscript{st}  and  2\textsuperscript{nd} round. Overall success is defined as follows: after the robot had stopped working, the participant grabs the parcel and puts it into the cardboard box, any time before the time out.}~\label{fig:figure_successV1-V4}
\end{figure}

\begin{figure}
  \centering
  \includegraphics[width=0.7\columnwidth]{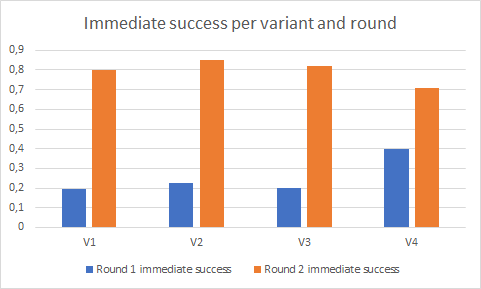}
  \caption{Comparison of successes rates (\%) between the 4 study conditions V1-V4. We distinguish immediate success within the 1\textsuperscript{st} and 2\textsuperscript{nd} round. Immediate success is defined as follows: immediately after the robot had stopped working, the participant grabs the parcel and puts it into the cardboard box.}~\label{fig:figure_immediate_successV1-V4}
\end{figure}

In fact, people were quite good in grasping what kind of help the robot needed in all conditions. See Figure \ref{fig:figure_successV1-V4} showing that over 90\% of the participants in conditions V1-V3 were already successful during the first round and 100\% were successful in the second round. The data also show a tendency that deictic gesturing of the robot arm supports humans to grasp that they should put the parcel in the box (94\% success rate in V2 round 1). Interestingly, participants in condition V4 were less successful (86\% success in round 1, and 91\% in round 2, which are both worse than the results in conditions V1-V3). As it seems, the combination of deictic gestures of the robot arm with the particular light signals, arrows running down the gripper, rather impairs than supports human understanding of the robot's intention, especially when already in doubt. 
For those, however, who were immediately successful, the combined signal in V4 seems to be a useful indicator, as two times more participants were immediately successful in V4 round 1 as compared to the first rounds in conditions V1-V3. See Figure \ref{fig:figure_immediate_successV1-V4} for illustration. 
Overall, only a small number of participants was immediately successful in first rounds, 20\% of the participants in conditions V1-V3, 40\% in condition V4. Whereas it were 80\% or more in round 2 of V1-V3. Here, V4 negatively stands out with only 71\% of the participants being immediately successful in round 2. 

\begin{table}
  \centering
  \begin{tabular}{|l|l|l|l||l|l|l|}
  \hline
 \small{\textbf{Condition}} & \multicolumn{3}{c}{\small{\textbf{Round 1}}} & \multicolumn{3}{c|}{\small{\textbf{Round 2}}}\\
   & \# & \#s & \#is & \# & \#s & \#is\\
  \hline
  V1 & 34 & 31 & 6 & 35 & 35 & 28 \\
  V2 &  33 & 31 & 7 & 33 & 33 & 28 \\
  V3 &  22 & 20 & 4 & 22 & 22 & 18 \\
  V4 & 35 & 30 & 12 & 34 & 31 & 22 \\
  \hline
\end{tabular}
  \caption{Number of participants per round and test condition; \# total number of persons in condition and round; \#s number of persons successful per condition and round; \#is number of persons immediately successful per condition and round.}
  \end{table}

\paragraph{Human behaviours after the robot had stopped working}

For the analysis of the behaviours the humans showed after the robot had stopped working, we concentrated on videos from participants who were not immediately successful in their first round per condition, and found the following behaviours in different frequencies:
\begin{itemize}\itemsep0pt
    \item Human looks at robot and waits (A) 
    \item Human tries to give the parcel to the robot (B) 
    \item Human rearranges the parcel (C) 
    \item Human tries to move the robot arm (D) 
    \item Human looks into the cardboard box before acting (E) 
    \item Human looks back and forth between robot and parcel (F) 
    \item Human searches for a signal in the environment (G) 
    \item Human looks at parcel arriving in front of the robot (H) 
    \item Human generally tries to intervene and take parcels off the conveyor belt already before the robot stops working (I) 
    \item Human already grabs the parcel before the robot stops (J) 
    \item Human gaze follows the parcel as it approaches on the conveyor belt (K) 
    \item Human makes gestures as if to say to the robot "go on working" (L) 
    \item Human tries previously learned system command gestures (thumbs up, palms up) which are for instance used to signal that the person is ready (M) 
    \item Human rearranges the parcels in the cardboard box (N) 
\end{itemize}

\begin{figure}
  \centering
  \includegraphics[width=0.9\columnwidth]{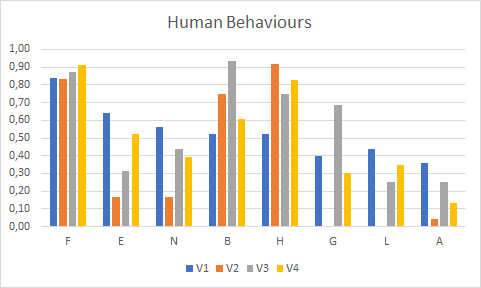}
  \caption{Frequent behaviours after robot had stopped working per first rounds of conditions V1-V4.}~\label{fig:figure_behaviours}
\end{figure}

The occurrence frequency of these signals differs greatly. In the following, we will discuss those with comparatively high occurrence in at least one of the conditions V1-V4. See Table \ref{tab:behaviours} for the frequency counts and Figure \ref{fig:figure_behaviours} for a graphical representation of the differences between the conditions. 

\begin{table}
    \centering
     \begin{tabular}{|l|l|l|l|l|}
     \hline
         \small{\textbf{Behaviour}} &
    \multicolumn{4}{c|}{\small{\textbf{Conditions}}} \\
    & V1 & V2 & V3 & V4 \\
    \hline
    F & 21 & 20 & 14 & 21 \\
    E & 16 & 4 & 5 & 12 \\
    N & 14 & 4 & 7 & 9 \\
    B & 13 & 18 & 15 & 14 \\
    H & 13 & 22 & 12 & 19 \\
    G & 10 & 0 & 11 & 7 \\
    L & 11 & 0 & 4 & 8 \\
    A & 9 &  1 & 4 & 3 \\
    \hline
    \small{\textbf{Persons in}} &  &  &  &  \\
    \small{\textbf{Condition}} & 25 & 24 & 16 & 23 \\
    \hline
    \end{tabular}
    \caption{Behaviours and occurrence frequencies per condition; only those trials were considered where the human was not immediately successful in the respective first rounds.}
    \label{tab:behaviours}
\end{table}

From Figure \ref{fig:figure_behaviours}, we see that there are some behaviours the participants show more often than others. The most frequent behaviour in all conditions was "the human looks back and forth between robot and parcel" (F), raging from 83\% in V2 to 91\% in V4. Moreover 94\% of the persons in V3, 75\% in V2, and still 61\% in V4 and 52\% in V1 try to give the parcel to the robot (B). Thus it seems that the light signal only (V3) was taken as a strong indicator for the humans to give the parcel to the robot. 92\% in V2, 83\% in V4 and 75\% in V3 look at the parcel at the time it arrives in front of the robot (H), which may be an indicator that specific signals such as deixis and or light as compared to no signal draw the human  attention to the object in focus. 69\% in V3 search for a signal in the environment, which may be due to a learning effect carrying over from other game sequences. 
In V1 and V4, more than half of the humans (64\% V1, 52\% V4) look into the cardboard box before they set an action (E). This was much less the case in V2 (17\%) and V3 (31\%). 56\% in V1 rearrange the parcels in the cardboard box (N). Which is only done by 17\% in V2. 
Overall, behaviours G, L, A, E and N were not at all or only rarely shown in V2, which we take as an indicator that robot deictic gestures are helpful.

\section{Conclusion and Future Work}

We presented a qualitative study where an industrial cobot in VR conducted a pick-and-place task and then signalled its human collaboration partner that it is their turn to pick-and-place objects. The signals were varied in four conditions: no signal (control group), robot arm gesture, light signal, combination of robot arm gesture and light signal. 

Between 20\% and 40\% of participants had immediate success in interpreting the signal of the robot in the first round. Between 70\% and 85\% were immediately successful in the second round. In general, more than 85\% of the participants were successful in the first round and 90\% up to 100\% in the respective second round. While the conditions with no signal, a robot arm gesture and a light gesture were relatively comparable, the fourth condition combining robot arm gesture and light signal stands out. While in the first round 40\% of participants were relatively successful (instead of 20\% in the other rounds), in the second round it was only 71\% (instead of 80\% or more in the other conditions). In order to gain more insights in this phenomenon, future research is needed. Another research question requiring further investigation is why some participants were immediately successful, while others needed more time or even failed. Different behaviours were exhibited if participants were not immediately successful, such as rearranging the objects, looking at the robot or the object, gesturing the robot to continue, trying to move the robot arm etc. We found evidence for the relevance of robot deictic gestures as source of information for the human collaboration partner what to do next. Moreover, we found evidence for a light signal projected on the gripper to be interpreted by the human as the robot's request to hand over the object in focus. However, this should be explored in further experiments and via interviews following the interaction, to gain insights in participants interpretations and expectations of industrial robots and their signals in specific application scenarios.

\section{Acknowledgments}
This work was supported by the Austrian Research Promotion Agency (FFG), Ideen Lab 4.0 project CoBot Studio (872590), and the Vienna Science and Technology Fund (WWTF), project "Human tutoring of robots in industry" (NXT19-005).


\balance{}

\bibliographystyle{SIGCHI-Reference-Format}
\bibliography{paper}

\end{document}